\documentclass[letterpaper, 10 pt, conference]{format/ieeeconf}
\IEEEoverridecommandlockouts
\usepackage{graphics} 
\usepackage{epsfig} 
\usepackage{times} 
\usepackage{amsmath} 
\usepackage{amssymb}  
\usepackage{todonotes}

\usepackage{amsfonts}
\usepackage{mathtools}
\usepackage{amsthm}
\usepackage{dsfont}
\usepackage{comment}
\usepackage{cite}
\usepackage{wrapfig}
\let\labelindent\relax
\usepackage{enumitem}
\usepackage{overpic}
\usepackage[font=footnotesize]{caption}
\usepackage[font=footnotesize]{subcaption}
\usepackage{algorithm}
\usepackage{algorithmic}
\usepackage[algo2e]{algorithm2e}
\usepackage{moreverb}
\usepackage[hyphens]{url}

\DeclareMathAlphabet{\mathcal}{OMS}{cmsy}{m}{n} 

\theoremstyle{definition}

\theoremstyle{remark}

\DeclarePairedDelimiterX{\norm}[1]{\lVert}{\rVert}{#1}

\newcommand{\rom}[1]{\uppercase\expandafter{\romannumeral #1\relax}}

{\end{list}}

\usepackage{xcolor}
\definecolor{DarkGreen}{rgb}{0,0.5,0.}

\newcommand{\troy}[1]{ {\color{orange}#1} }

\title{\LARGE \bf Data-Efficient Learning of High-Quality Controls for\\ Kinodynamic Planning used in Vehicular Navigation}

\author{Seth Karten, Aravind Sivaramakrishnan, Edgar Granados, Troy McMahon, Kostas E. Bekris%
\thanks{\textit{Machine Learning for Motion Planning (MLMP) Workshop at ICRA 2021}, Xi'an, China. The authors are with the Deparment of Computer Science, Rutgers University, NJ, USA. Email: {\tt\small { \{sak295, as2578, eg585, tm799, kb572\}}@rutgers.edu}}%
}

\begin{document}
\maketitle
\thispagestyle{empty}
\pagestyle{empty}

\begin{abstract}
This paper aims to improve the path quality and computational efficiency of kinodynamic planners used for vehicular systems. It proposes a learning framework for identifying promising controls during the expansion process of sampling-based motion planners for systems with dynamics. Offline, the learning process is trained to return the highest-quality control that reaches a local goal state (i.e., a waypoint) in the absence of obstacles from an input difference vector between its current state and a local goal state. The data generation scheme provides bounds on the target dispersion and uses state space pruning to ensure high-quality controls.  By focusing on the system's dynamics, this process is data efficient and takes place once for a dynamical system, so that it can be used for different environments with modular expansion functions. This work integrates the proposed learning process with a) an exploratory expansion function that generates waypoints with biased coverage over the reachable space, and b) proposes an exploitative expansion function for mobile robots, which generates waypoints using medial axis information. This paper evaluates the learning process and the corresponding planners for a first and second-order differential drive systems. The results show that the proposed integration of learning and planning can produce better quality paths than kinodynamic planning with random controls in fewer iterations and computation time. 
\end{abstract}

\section{Introduction}\label{sec:intro}


There has been an increased focus on developing motion planners that identify high-quality (i.e., low-cost) paths.  In the context of sampling-based planners, this has been exemplified by the development of the asymptotically optimal PRM*, RRT* \cite{KF-2011} algorithms as extensions of the original, probabilistically complete PRM \cite{KSLO-1996} and RRT \cite{L-1998}.  A limitation of PRM* and RRT* is that they require a steering function for connecting two states with a trajectory. For many systems with kinodynamic constraints, such a function is not readily available.  This has lead to the development of tree-based planners similar to RRT, which do not require a steering function  and aim towards asymptotic optimality \cite{Hauser_2016,kleinbort2019refined,westbrookanytime}. Examples of these methods include prior work by some of the authors on SST \cite{Li:2016} and its informed variant DIRT \cite{LB-DIRT}. These methods use only forward propagation of random controls. This is desirable for the analysis of their properties but in practice random controls result in slow convergence to high-quality trajectories.  


\begin{figure}[t]
  \begin{center}
    \includegraphics[width=.4\textwidth]{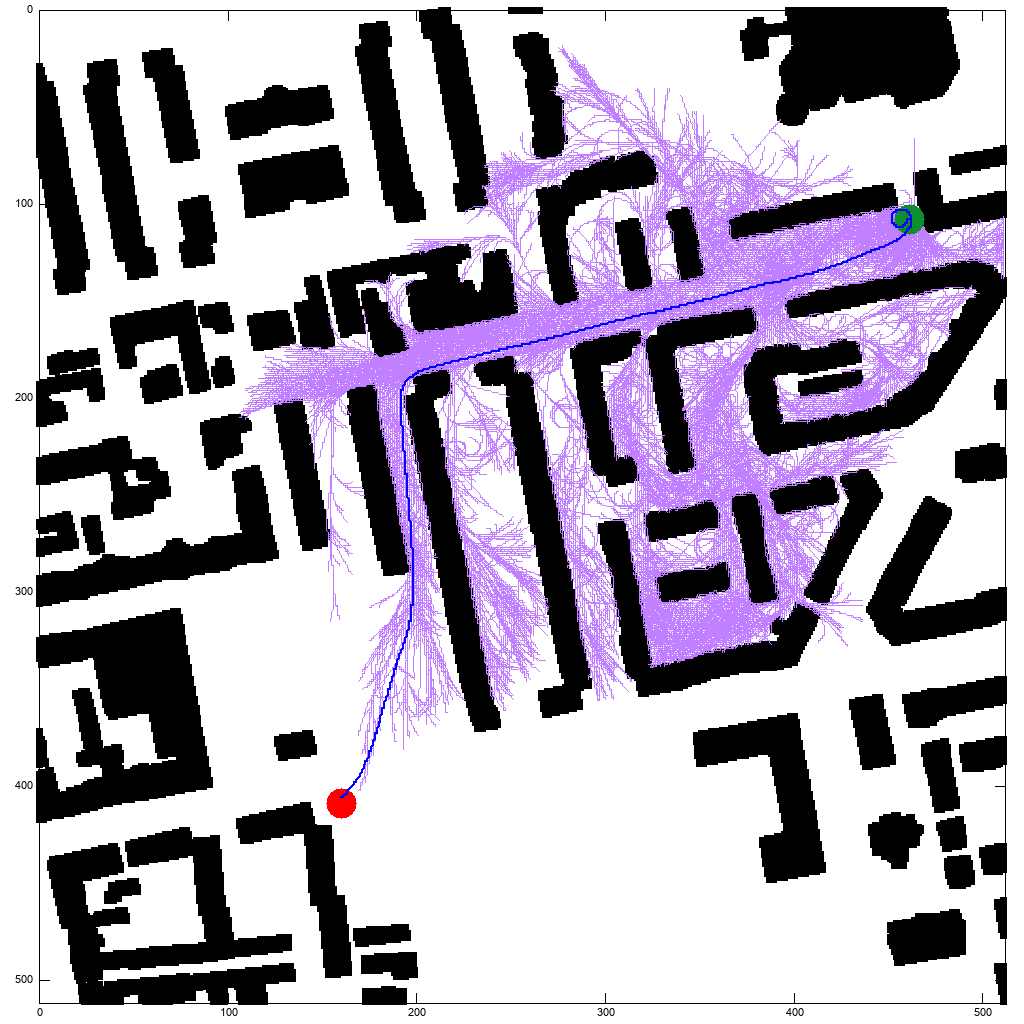}
    \caption{Application of the proposed learned controller with the medial axis expansion to a second-order mobile robot navigating the Berlin map.}
    \label{fig:roadmap}
  \end{center}
\end{figure}


This work proposes learning high-quality controls, which acts as a surrogate for the dynamics of the system, and integrating the learned model with asymptotically optimal kinodynamic planners to improve their practical performance while maintaining their desirable properties.  This method is built so as to fit the specifications of a propagation function in a sampling-based planner, such as DIRT \cite{LB-DIRT}, which is used in the accompanying experiments.  The approach consists of a supervised process, which operates over a difference vector between a current state and a local target state, i.e., a waypoint towards making progress to a global goal. It's tasked to return a constant control and a corresponding duration of propagation. The objective is for this control and duration to produce a trajectory that brings the robot to a state that minimizes distance to the waypoint state.  This work describes an efficient data collection method for this supervised learning challenge, which provides bounds on the target dispersion and control quality. 

In addition to the learned model, this paper proposes both exploitative and exploratory propagation functions for the kinodynamic planners, which generate waypoints to be passed to the learned controller to return the best control. The benefit of this two-step approach is that the learning process has to only deal with the system's dynamics and needs to be trained once for a specific robot. The waypoint selection and the planning process are responsible for dealing with different environments and selecting a direction of progress towards the global goal.


The accompanying simulated experiments use the learned controller and the different propagation strategies in combination with the DIRT planner. The evaluation uses a first-order and second-order differential drive system across different environments. It shows that high-quality controls in response to desirable waypoints can be learned efficiently without high data requirements. When integrated with the planner, the learning-based approach results in very low cost solutions that are achieved in less time and fewer iterations relative to traditional random propagation approaches.

\section{Related Work}
\label{sec:related_work}

There have been multiple applications of machine learning in motion planning. For instance, methods learn a sampling distribution to bias sampling along a desired trajectory for a specific planning problem or environment \cite{IHP-2018,qureshi2018deeply,Wang-2020}. Work has also been performed to find end-to-end solutions from expert trajectories \cite{Pfeiffer_2017,QBY-2018}. These methods require an expensive data-driven approach gathering multiple solutions to supervise the corresponding. Furthermore, they are complementary to the approach described here as they focus on the sampling process of a planner and not control propagation. 

Continual learning \cite{quereshi2020motion} is able to significantly reduce the number of expert solutions necessary, but has only been applied to solution paths rather than trajectories that provide solutions with controls. Methods have used reinforcement learning to approach planning with dynamics by using a learned model to generate controls in PRM-based \cite{forftfd-2018} and RRT-based \cite{Chiang_2019} approaches, but require large amounts of data to learn the policy and tune the reward parameters.  

Conditional variational autoencoders (CVAE) \cite{IHP-2018} have been used to learn sample distributions from demonstration.  Samples are then generated by conditioning the CVAE on problem variables such as the start and the goal.  This method differs from ours in that it generates samples rather then controls or trajectories and consequently is more suited for traditional motion planning then for kinodynamic planning.


Another RL approach proposes a motion planner for aerial robots with suspended loads \cite{FPCFT-2017}. It applies reinforcement learning to learn a model for generating trajectories that satisfy the dynamic constraints introduced by the suspended loads.  
\emph{Continuous action fitted value iteration} (CAFVI) \cite{faust-acta-14} uses reinforcement learning in combination with sampling based motion planning for affine nonlinear systems where the dynamics are unknown or intractable. 
The Dynamics-Aware Discovery of Skills (DADS) algorithm \cite{sharma2019dynamics,sharma2020emergent} uses unsupervised learning to learn robotic \emph{skills}, including gaits for walking robots.
These methods primarily address the problem of generating immediate, constraint satisfying motions whereas our work focuses on generating motions that make progress towards a long-term goal state.

Another approach, Near-Optimal RRT (NoD-RRT) \cite{LCLX-2018}, uses a neural network to estimate cost and can be applied to problems with nonlinear kinodynamic constraints as a heuristic to estimate the cost between states. Nonlinear parametric model with constant-time inference \cite{pa-2015} have also been used to approximate distances for wheeled robots.
While such methods are useful for deciding what nodes to expand, they still rely on forward propagation to do the expansion.  In comparison, our model is able to select promising controls, which allows us to generate better quality trajectories during expansion. 

There are also many traditional approaches for biasing the operation of sampling-based planners. For instance, one can use the environment's medial axis \cite{Xu-1992-15885} to generate samples along the medial axis of the configuration space \cite{MAPRM,UMAPRM} or generate an RRT along the medial axis \cite{MARRT}. 
These methods do not take the dynamics of the system into account or provide asymptotic optimality. They have motivated, however, the development of the exploitative propagation strategy which generates waypoint states using medial axis information, then passes them to the proposed controller. 

Furthermore, one way to improve the convergence rate of sampling-based planners is to consider an "RRT-blossom" expansion of controls \cite{KP-2006}. It propagates multiple controls per iteration, to better approximate the reachability set. In combination with a heuristic, the most promising control can be selected more frequently to give a high-quality solution earlier. The issue, however, is that this approach performs multiple propagations per iteration, which is expensive. This work uses this as a comparison point for showing the benefits of the learned controller against even the curated propagation of multiple controls as in the "RRT-blossom" approach.

\section{Problem Setup and Notation}  \label{sec:problem_statement}
Consider a system with state space $\mathbb{X}$ and control space $\mathbb{U}$, governed by the dynamics $\dot{x} = f(x,u)$ (where $x \in \mathbb{X}, u \in \mathbb{U}$). The objective is to learn a controller $\hat{f}(x,\delta x)$, which returns a control $u$ and duration  $\delta t$, so that if the control $u$ is propagated from state $x$ so that the system can achieve a \textit{desired} change in the state space $\delta x$. 

This work uses a first-order and a second-order differential drive system as motivating examples. The state space for the first-order system is $\{x,y,\theta\}$, and its control space is $\{v_L,v_R\}$. The state space for the second-order system is given by $\{x,y,\theta,v_L,v_R\}$, and its control space is $\{u_L,u_R\}$. 

Define a distance function $d$ that compares independently across noncomparable degrees of freedom, which will be used in pruning operation of sampling data. In the first-order system, the Euclidean distance function is $d_E$, $d_E(s, s\prime) = \sqrt{(s.x-s\prime.x)^2+(s.y-s\prime.y)^2}$, and a rotational distance function $d_R$, $d_R(s,s\prime)= (s.\theta-s\prime.\theta) \% \pi$. For pruning purposes,  $d$ is under some threshold $\epsilon$ if $d_E < \epsilon_E$ and $d_R < \epsilon_R$. For the second-order system, similarly define the distance function with the addition of a velocity distance function $d_V$, $d_V(s,s\prime) = \sqrt{(s.v_l-s\prime.v_l)^2+(s.v_r-s\prime.v_r)^2}$. For this system, there is also the constraint that $d < \epsilon$ such that $d_V < \epsilon_V$. For more complicated systems, supplementary distance functions can be defined similarly.

Define the reachable set as the region of the state space, $R_{x,\delta t}$ that the robot can reach from state $x$ in timestep $\delta t$ given constant controls. Formally $R_{x,\delta t}$ is the set of all states $x'$ such that there exists a constant-control trajectory $\tau_{x,x'}$ that starts at $x$, ends at $x'$ and requires at most time $\delta t$.
\begin{equation*}
    \begin{split}
        \mathbb{R}_{x,\delta_t} = &\{ x' \vert x' = \texttt{Propagate}(x,(u,\delta_t))\\
        &\forall u \in \mathbb{U}, \ \forall \delta t \in [t_{MIN},t_{MAX}]\}
    \end{split}
\end{equation*}

Define the medial axis of a workspace as a set of points $p$ for which there are two or more closest points on obstacles.
\begin{equation*}
    \begin{split}
        \mathbb{MA}(env) = & \{ p \in env|free(p) \wedge \exists \{(p_1,p_2| p_1 \ne p_2 \wedge \\
& dist(p,p_1)=dist(p,p_2)\wedge obs(p_1) \wedge obs(p2) \wedge \\
& \not \exists \{p_3 | obs(p_3) \wedge dist(p,p_3)<dist(p,p_1)\}\}\}
    \end{split}
\end{equation*}

\noindent where free(p) indicates that p is a point in the free region of workspace, obs(p) indicates that p is a point on an obstacle and dist($p$,$p_1$) is the Euclidean distance between $p$ and $p_1$. 

\section{Proposed Method}\label{sec:method}

\begin{algorithm}[tb]
\SetAlgoLined
\begin{algorithmic}[1]
    \STATE {$u_j \gets u_j^{min}$ s.t. $j\neq i$.} 
    \STATE Propagate the plan $([u_i, \{u_j\}],\delta t)$ from $x_0$  to obtain final state $x_1$.
    \STATE $N_{controls} \gets N_{min} - 1$.
    \REPEAT
        \STATE Increment $N_{controls}$.
        \STATE  $u_i \gets u_i$ + ($u_i^{max}$ - $u_i^{min})/N_{controls}.$ 
        \STATE Propagate the plan $([u_i, \{u_j\}],\delta t)$ from $x_0$  to obtain final state $x_2$.
    \UNTIL{$d_E(x_1,x_2) < \epsilon_E$}
    \RETURN $N_{controls}$
\end{algorithmic}
 \caption{\texttt{CTRL-RESOLUTION}$(x_0,u_i,\epsilon_E,\delta t$)}
 \label{alg:ctrl-resolution}
\end{algorithm}

This section describes the learned controller, which given as input the difference between the current state and a local goal state, outputs a control and propagation duration, which results in a trajectory that minimizes the distance to the local goal state.  The model is trained in an obstacle-free environment in order to capture the dynamics of the system. The training data are tailored to ensure adequate coverage over the reachable state space and to minimize the time to reach any reachable state. The data collection process takes advantage of symmetry by transforming at least a subset of the state $x$ to the origin. For the first-order system, this is applicable to the entire state, since the reachable set relative to the initial state is the same for all initial states. For the second-order system, the reachability set changes depending on the initial state's velocities. In this case, the $\{x,y,\theta\}$ DoFs of the state space are transformed to the origin, while the remaining velocity parameters are treated as free variables for which there is a need to sample additional data. Thus, $\{\delta x, \delta y, \delta \theta\}$ are used as the inputs for these degrees of freedom. For other degrees of freedom that do no have an obvious difference, such as the velocity states for the 2nd order vehicle, the input is the initial and waypoint velocities.

The input provided to the learned controller when used online may be incorporated in an exploratory or exploitative manner. A learned controller used with an exploratory input needs to build a set of high-quality controls to find low-cost solutions during motion planning. The sample-based expansion samples random inputs since any branches added to the search tree will be high-quality. A learned controller used with an exploitative input must ensure that it can target any reachable state. A proposed medial axis expansion queries a medial axis data structure to obtain a waypoint that indicate the direction the robot needs to move to reach the goal while avoiding collisions. In practice, experiments indicate that the learned controller balances its exploration ability and exploitative capabilities. 

Both the exploratory and exploitative expansion methods can be used by planners, such as RRT and DIRT. This work uses these expansion functions as part of the DIRT planner with blossom expansion.  In the context of the blossom expansion, the first propagation is exploitative, i.e., towards the goal according to the medial axis, and the remaining expansions are exploratory. The methods that used the learned controllers only use learned controls in the blossom to ensure that the search tree is built with higher quality controls.

\begin{algorithm}[t]
\SetAlgoLined
\begin{algorithmic}[1]
    \STATE $U^{ctrls}$ $\gets$ \texttt{DISCRETIZE-CTRLS}($\epsilon$)
    \STATE \texttt{data} $\gets$ \texttt{PROPAGATE-AND-PRUNE}($U^{ctrls}, \epsilon$)
    \STATE \texttt{data} $\gets$ \texttt{SAMPLE-PROP-PRUNE}(\texttt{data}, $N, \epsilon$)
    \RETURN \texttt{data}
\end{algorithmic}
\caption{\texttt{GENERATE-CTRL-DATA}($\epsilon$)}
\label{alg:ctrl-data}
\end{algorithm}

\subsection{Dataset Construction for Learning Controls}
\label{sec:dataset}

In this section we discuss the dataset we use to train the controller. There are two desirable properties for this dataset: \textbf{target dispersion} and \textbf{control quality}. By ensuring that the constructed dataset satisfies these properties, the learned controller will satisfy them with some bounded error. 

For a distance function $d$, the target dispersion imposes the following condition: the distance between any state $s$ in the dataset and its nearest neighbor $s_n$ is less than $\epsilon$. This ensures that the dataset adequately covers the entire reachable set of the system for a given initial state.

The second property is important in the context of motion planning. For a given initial state, it is possible for multiple control sequences, i.e., controls propagated for different durations, to reach a similar $\delta x$ (or its $\epsilon$-neighborhood). 
For a given $\delta x$, it is desirable that the dataset should only contain the control sequence with the lowest cost, which corresponds to the control propagation duration. For any two states $s$ and $s\prime$, such that $d(s,s\prime) \leq \epsilon$, only the state with the lower cost is added to the dataset.

\begin{figure}[tb]
  \begin{center}
  \vspace{-4mm}
    \includegraphics[width=.4\textwidth]{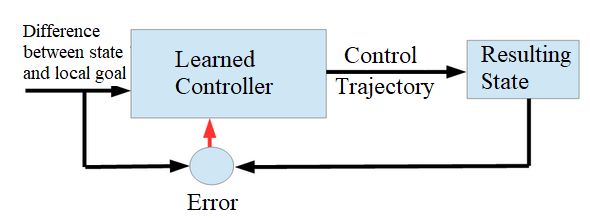}
    \includegraphics[width=.4\textwidth]{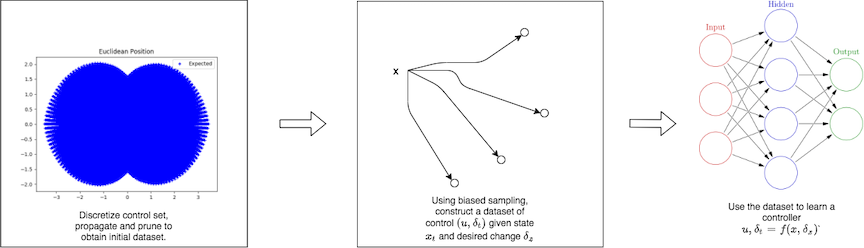}
    \caption{The controller is trained to return the shortest duration controls that can achieve the difference between the initial state and a local goal state.  The difference between these two states is the error, which is used to update the model.}
    \label{fig:training_worker}
     \vspace{-3mm}
  \end{center}
\end{figure}

For any system, the dataset is stored in a GNN nearest neighbor data structure, using euclidean distance as the key.
Initially, the procedure sets a target discretization dispersion value $\epsilon_E$ for the euclidean state $(x,y)$. Ensuring $d(s,s_n) < \epsilon_E$ is largely governed by the discretization chosen for partitioning the range of durations as well as the control space bounds. Too fine discretization will waste a lot of propagation steps, and too coarse will result in high dispersion in the reachable state space. The propagation duration must be given special consideration, since controls propagated for longer durations will end up in states that are farther apart. In order to adequately cover the reachable region of the state space, we use Algorithm~\ref{alg:ctrl-resolution} to approximate the level of discretization required for the given initial state conditions (if any) and the current propagation time.

As each control is added to the dataset, the approach prunes samples to get an optimal dataset. Given two states that satisfy $s,s\prime$ $d(s,s\prime) < \epsilon$, it retains the datapoint with the shortest duration. There are at most $k$ distance parameters to tune for pruning in a state space with $k$ variables. The parameters are hand-tuned and evaluated given training error and performance during planning.

After the discretization, the procedure uses random sampling in the control space to further improve the quality of the dataset by comparing each datapoint to at least $N$ other datapoints. The random sampling fills in any gaps in the reachable state space that were created by discretization.

There is a trade-off between the degree of pruning and the reachability of the learned controller. More pruning will ensure that the learned controls are faster, leading to lower cost planning solutions. However, there may be unreachable states that can lead to better solutions, e.g. the learned controller may not be able to achieve tight turns. In contrast, less pruning will lead to have more reachable states. As the pruning constraint is relaxed, each datapoint will be compared to fewer alternatives, leading to slower controls on average. But they allow more precise turning or specialized maneuvers. Tuning this balance is easier for simple systems, but in the more complex cases, the pruning can be more lenient and still ensure high-quality solutions by biasing the inputs in the manner described in Sections \eqref{sec:reachability} and \eqref{sec:MA}.

\subsection{Sample-Based Expansion}\label{sec:reachability}
The first proposed variant for using the learned controller inside a kinodynamic planner during propagation is an exploratory one. It generates waypoints by randomly sampling in the robot's reachable state space. Then, the planner's \texttt{expand} function uses the learned controller to generate controls that move towards the waypoint.  

In particular, the sample-based expansion function randomly generates a waypoint in the reachable set given the robot's current state. Then, it uses the learned model to generate a control that takes it towards the waypoint. Biased sampling in the reachable set provides coverage over the reachable region while biasing towards higher quality controls. This leads to faster exploration. In contrast, the typical expansion of states using random controls does not produce a uniform distribution and results in samples that are clustered around the center of reachable space.

\begin{figure}[tb]
  \begin{center}
    \includegraphics[width=0.15\textwidth]{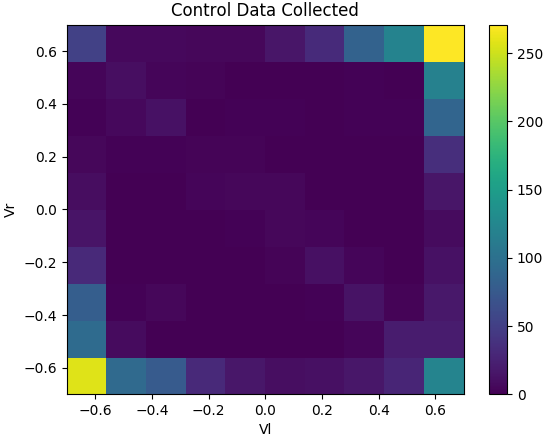}
    \includegraphics[width=0.15\textwidth]{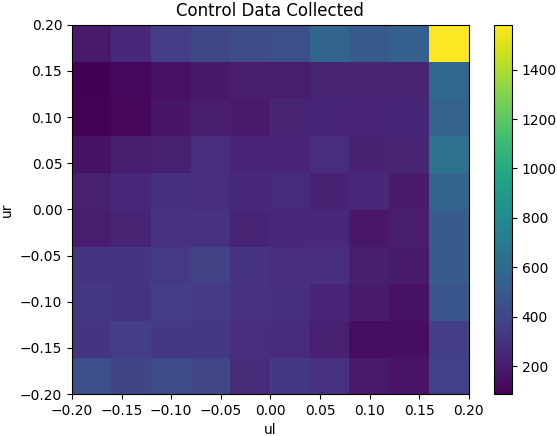}
    \includegraphics[width=0.15\textwidth]{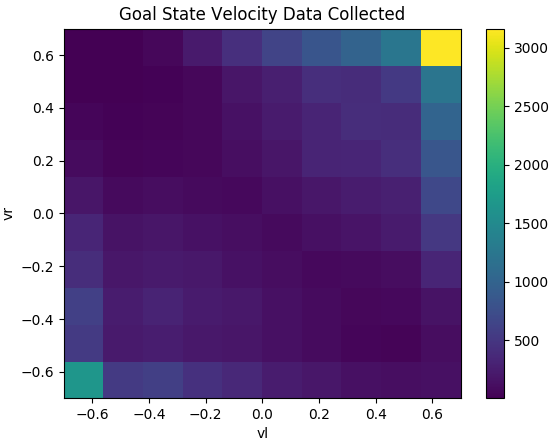}
    \caption{Control data histogram collected for the 1$^{st}$ order system with $v_l,v_r$ (left) and 2$^{nd}$ order system with controls $u_l,u_r$ (middle). The dataset favors bang-bang solutions. (right) Frequency of the velocities $v_l,v_r$ for the local goal states used in training. The local goal state for the velocities of the 2$^{nd}$ order system is sampled from this distribution.}
    \label{fig:velDist}
    \vspace{-6mm}
  \end{center}
\end{figure}

%

For more complicated systems, biased sampling for the additional degrees of freedom in the state space, such as $v_l$ and $v_r$ in the second-order system, can be used. Since the collected dataset has already pruned low-quality local goal states, a 2D histogram of the initial velocities in the dataset (see figure \ref{fig:velDist}) can be transformed into a biased sampling distribution. For each bucket of the 2D histogram, there are a number of datapoints. A uniform random variable $v$ is mapped to the range [0, $N_V$], $N_V$ the total number of examples in the velocity data. From the output of $v$, one can subtract the number of datapoints in each bucket until the value is $\leq 0$. This bucket can now be uniformly sampled to produce the local goal velocity.


The produced local goal is fed to the learned controller. Its output is the control and propagation duration to be used by the planner at the current tree node.

\begin{figure}[tb]
  \begin{center}
    \includegraphics[width=0.15\textwidth]{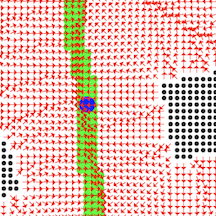}
    \includegraphics[width=0.15\textwidth]{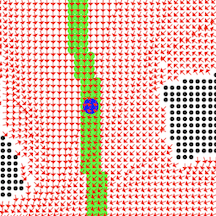}
    \includegraphics[width=0.15\textwidth]{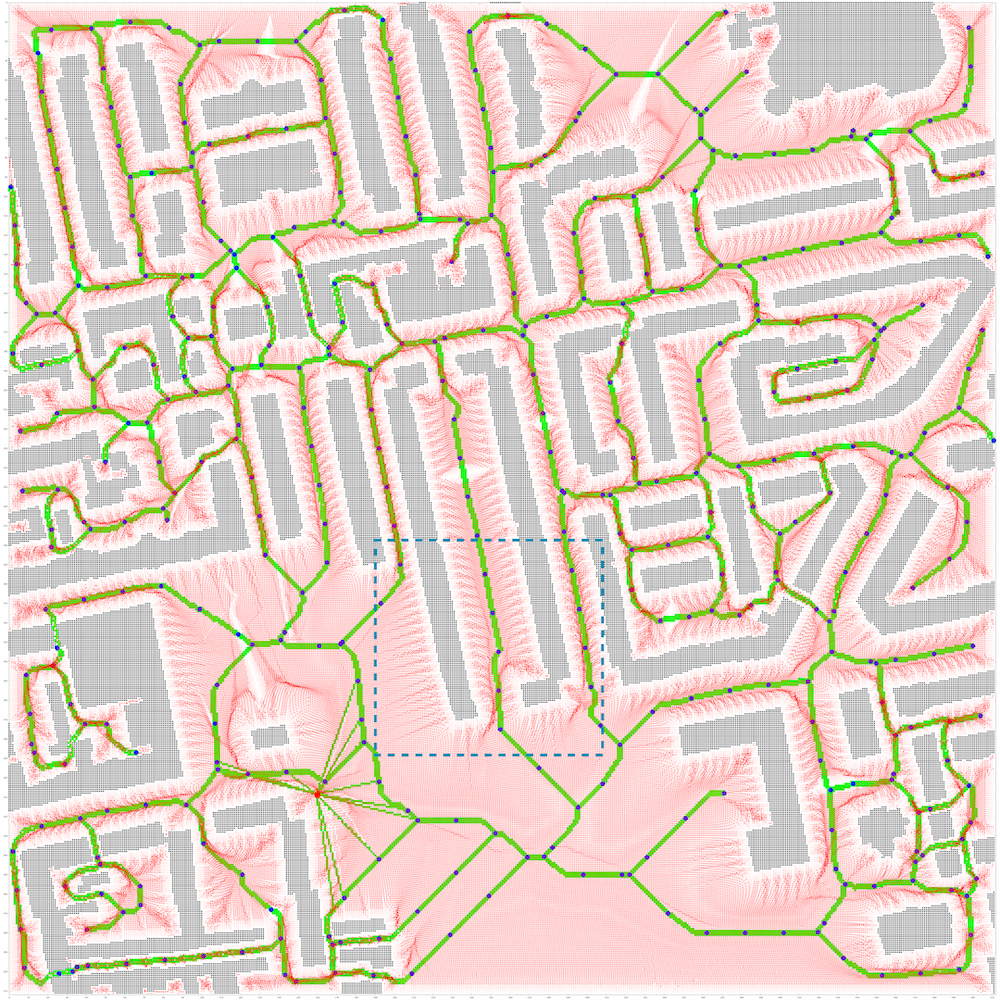}
    \caption{Medial axis and vector fields for the Berlin environment: (left) Close-up of repulsive normalized vector field (red) and medial axis (green/blue) (middle) Close-up of attractive normalized vector field  (right) \textit{Integrated} vector field. Shown in full here, each vector is computed online as requested. Dotted blue rectangle shows close-up region.}
    \vspace{-.1in}
    \label{fig:ma_planner}
  \end{center}
\end{figure}

\subsection{Medial Axis Expansion}\label{sec:MA}
Next, an exploitative variation of the planner is proposed. It uses a medial axis of the workspace to compute waypoints that are taken as input by the learned model. The objective of the medial axis is to reduce exploration time while keeping good path quality. The medial axis is used to find a vector from the robot's current position $s$ to an ideal next position $s\prime$. Using medial axis points as local goals is advantageous because points on the medial axis are as far away from obstacles as possible.
Intuitively, the closer a robot is to the medial axis, the higher it velocity can be without getting into an inevitable collision state (ICS). Additionally, the medial axis can be use to compute a direction that will get the robot closer to the goal.

The medial axis pipeline is illustrated in Figure \ref{fig:ma_planner}.
The first step is done offline. It consists of generating a repulsive vector field $U_{rep}$ and computing the workspace medial axis of a specific environment. Then, the goal is added to the medial axis and an attractive vector field $U_{att}$ is computed. Each vector $v_{i}$ on $U_{att}$ aims to a point $p_{i}$ such that: $p_{i}$ is on the medial axis, $v_{i}$ has direct line of sight (with a clearance) to $p_{i}$ and the cost-to-go of $p_{i}$ is minimum. Online, both vector fields are used compute a normalized \textit{integrated} vector: $U_{int}(x)=w * U_{rep}(x) + (1-w) * U_{att}(x)$. The vector is then \textit{extended} such that its magnitude either complies with the reachability of the robot or it reaches the medial axis. 

The medial axis is used to determine the $\delta x, \delta y,$ and $\delta \theta$ for learned controller input. The $\delta x$ and $\delta y$ are found from taking the difference between the current position $s$ and ideal next position $s\prime$. The $\delta \theta$ is found from difference between the current rotation and the angle created from $s\prime$ to its ideal next position $s\prime\prime$. Other additional degrees are freedom are sampled in the manner used in the sampling-based expansion (section \ref{sec:reachability}).

\begin{figure}[tb]
  \begin{center}
    \includegraphics[width=0.23\textwidth]{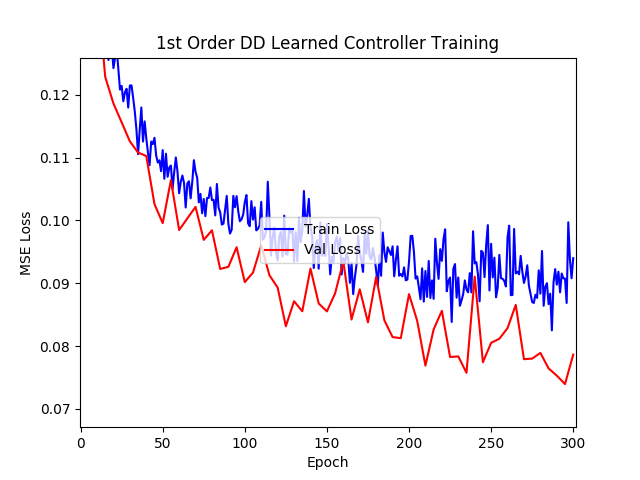}
    \includegraphics[width=0.23\textwidth]{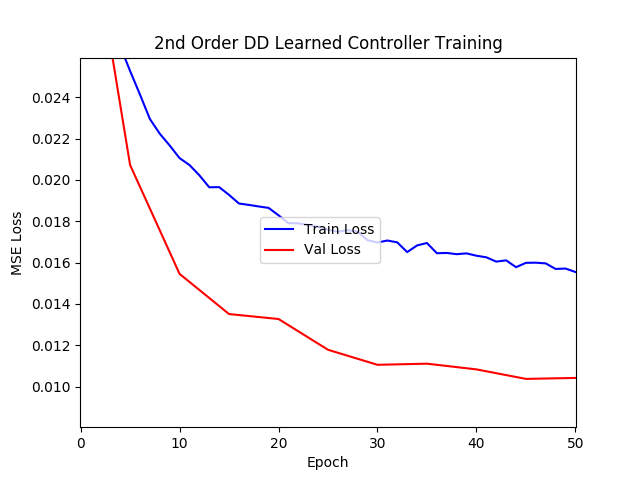}
        \vspace{-.05in}
    \caption{Training results for the $1^{st}$ (left) and $2^{nd}$ order system (right).}
        \vspace{-.1in}
    \label{fig:training}
  \end{center}
\end{figure}

\subsection{Model Training Setup}

The network architecture uses a sequence of fully-connected layers with 1D batch normalization and a \emph{tanh} activation function, which allows the network to execute bang-bang controls for inputs outside of the reachability range. The euclidean ($\delta x,\delta y$) position is mapped to polar coordinates ($\delta r, \delta phi$). All inputs and outputs are further mapped to [-1,1]. Online, they are mapped back to the state and control space bounds, respectively. The approach tunes an error function that combines a weighted mean-squared error over the control-duration space and the propagated state space. A separate neural network is trained to learn the forward propagation so that the gradient signal can be backpropagated from the propagated state space loss. The number of network parameters was limited to below 2000 for each system to ensure fast inference time. 

\begin{figure}[tb]
  \begin{center}
  \vspace{-5mm}
    \includegraphics[width=0.15\textwidth]{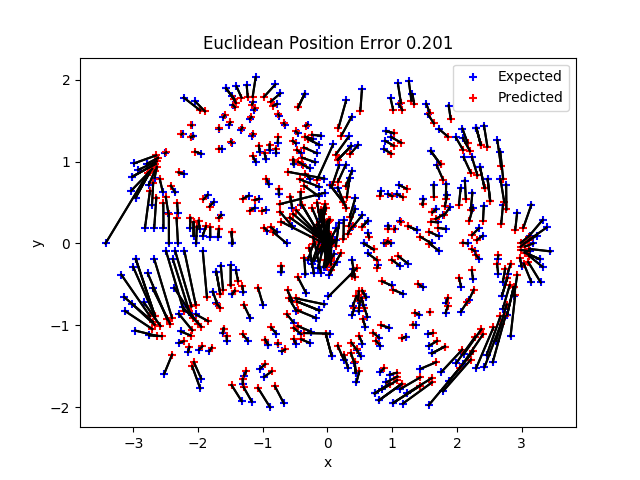}
    \includegraphics[width=0.15\textwidth]{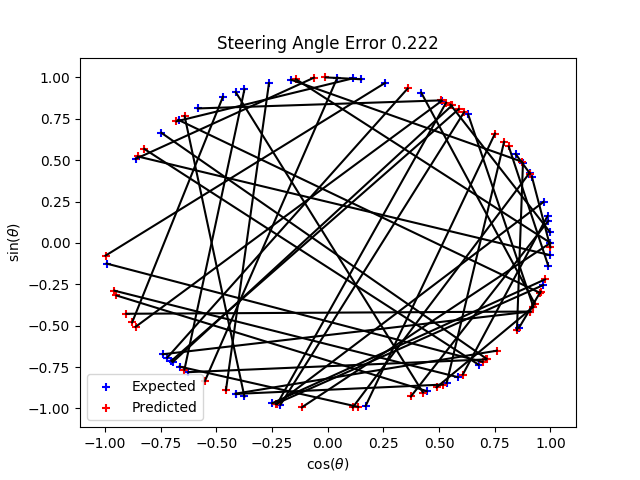}\\
    \includegraphics[width=0.15\textwidth]{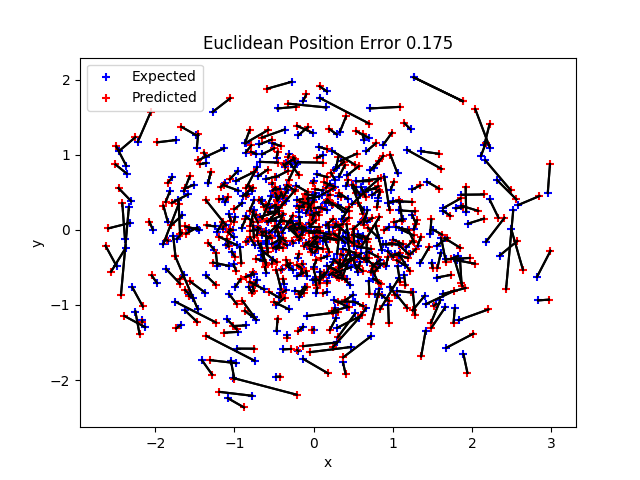}
    \includegraphics[width=0.15\textwidth]{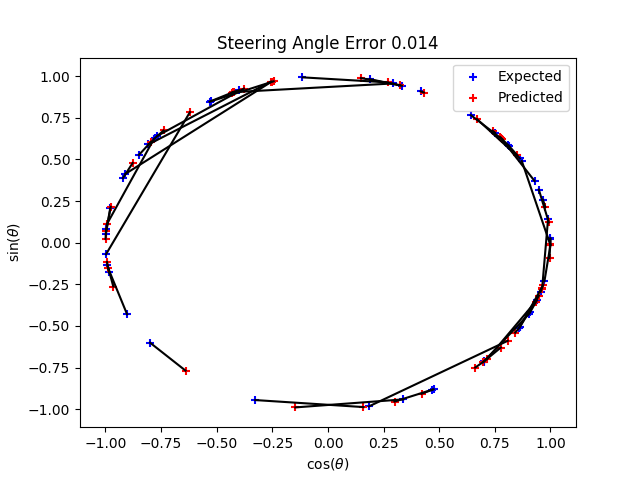}
    \includegraphics[width=0.15\textwidth]{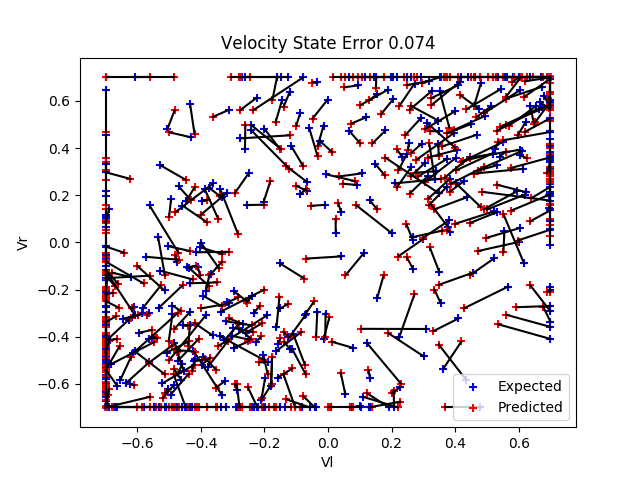}
        \vspace{-.05in}
    \caption{Comparison of 500 random predicted results vs actual results for controls generated using the trained model in a $1^{st}$ order system (top) and a $2^{nd}$ order system (bottom).  The blue dots indicate the final positions for each trajectory and the red dots indicate the predicted final state by the model. Lines connect corresponding pairs of actual and predicted states for euclidean, rotational, and, for the 2nd order system, velocity terms. }
    \vspace{-3mm}
    \label{fig:connected_reachabilityr}
  \end{center}
\end{figure}

\begin{figure}[t]
    \centering
         \includegraphics[width=.15\textwidth]{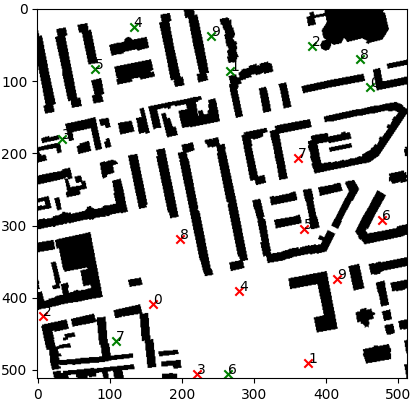}
         \includegraphics[width=.15\textwidth]{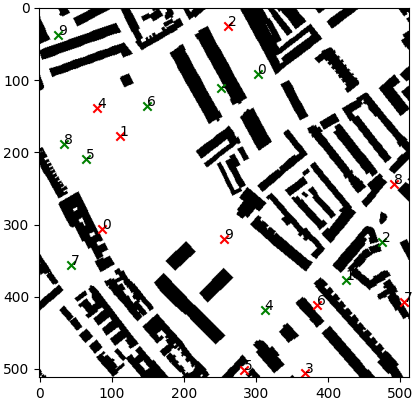}
         \includegraphics[width=.15\textwidth]{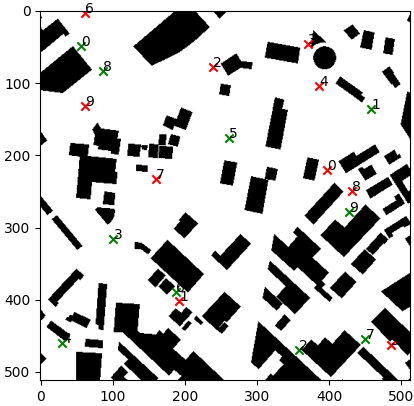} 
    \caption{Environments:  City maps from Berlin (left), London (center) and Boston (right). 
    }
    \label{fig:city_environments}
\end{figure}

\begin{figure*}
\centering
\begin{subfigure}[b]{\textwidth}
          \includegraphics[width=.245\textwidth]{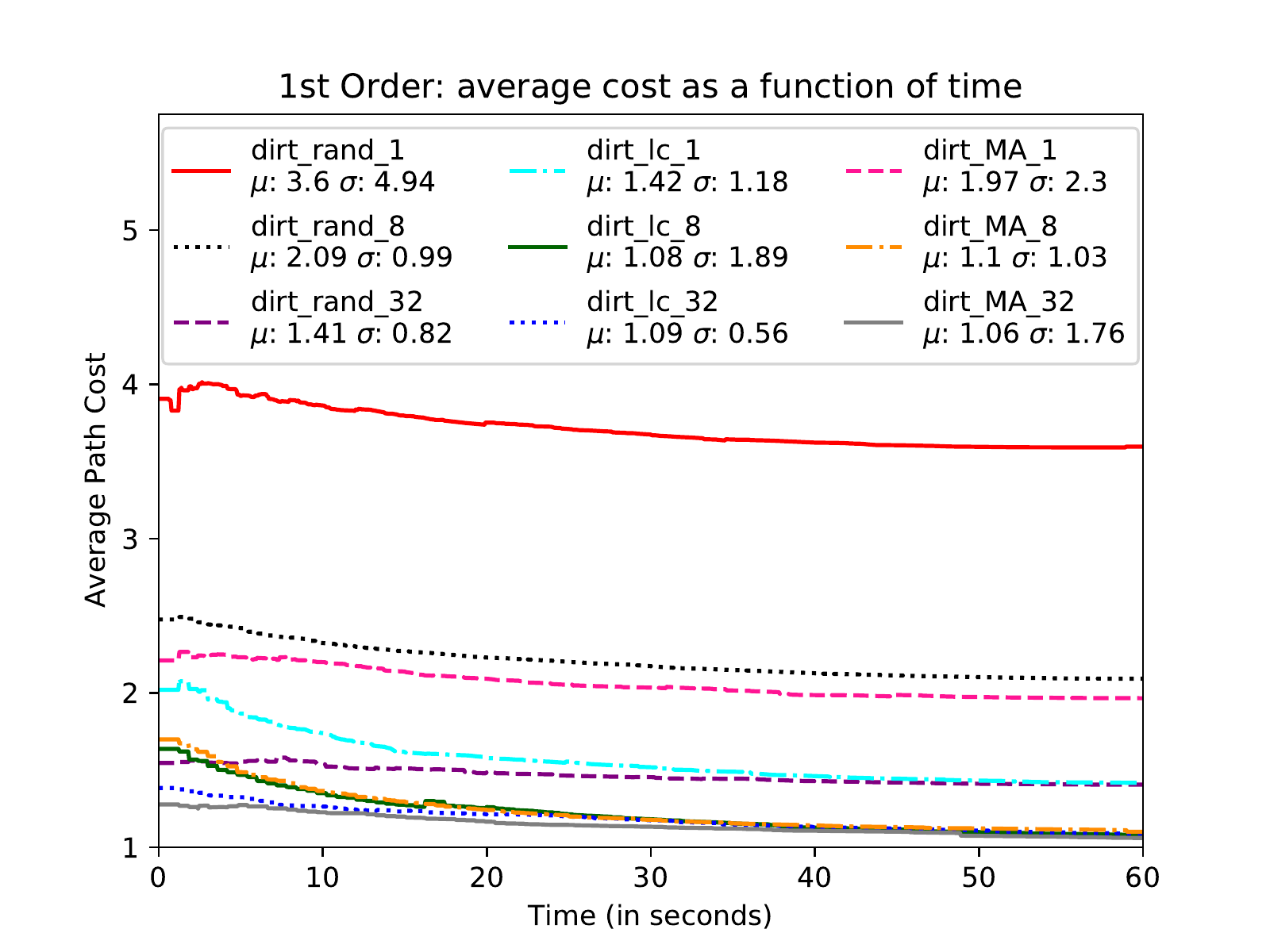}
         \includegraphics[width=.245\textwidth]{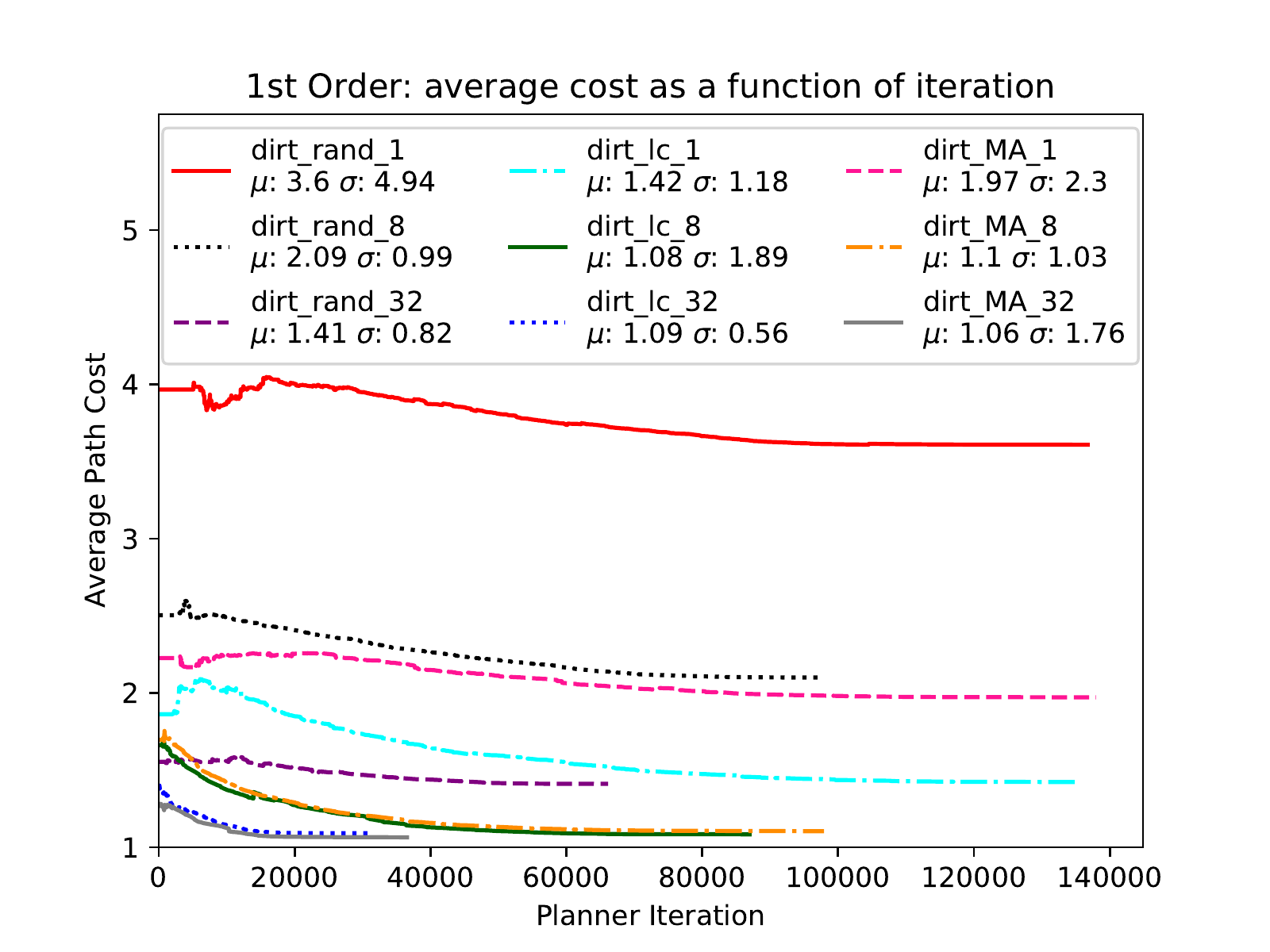}
         \includegraphics[width=.245\textwidth]{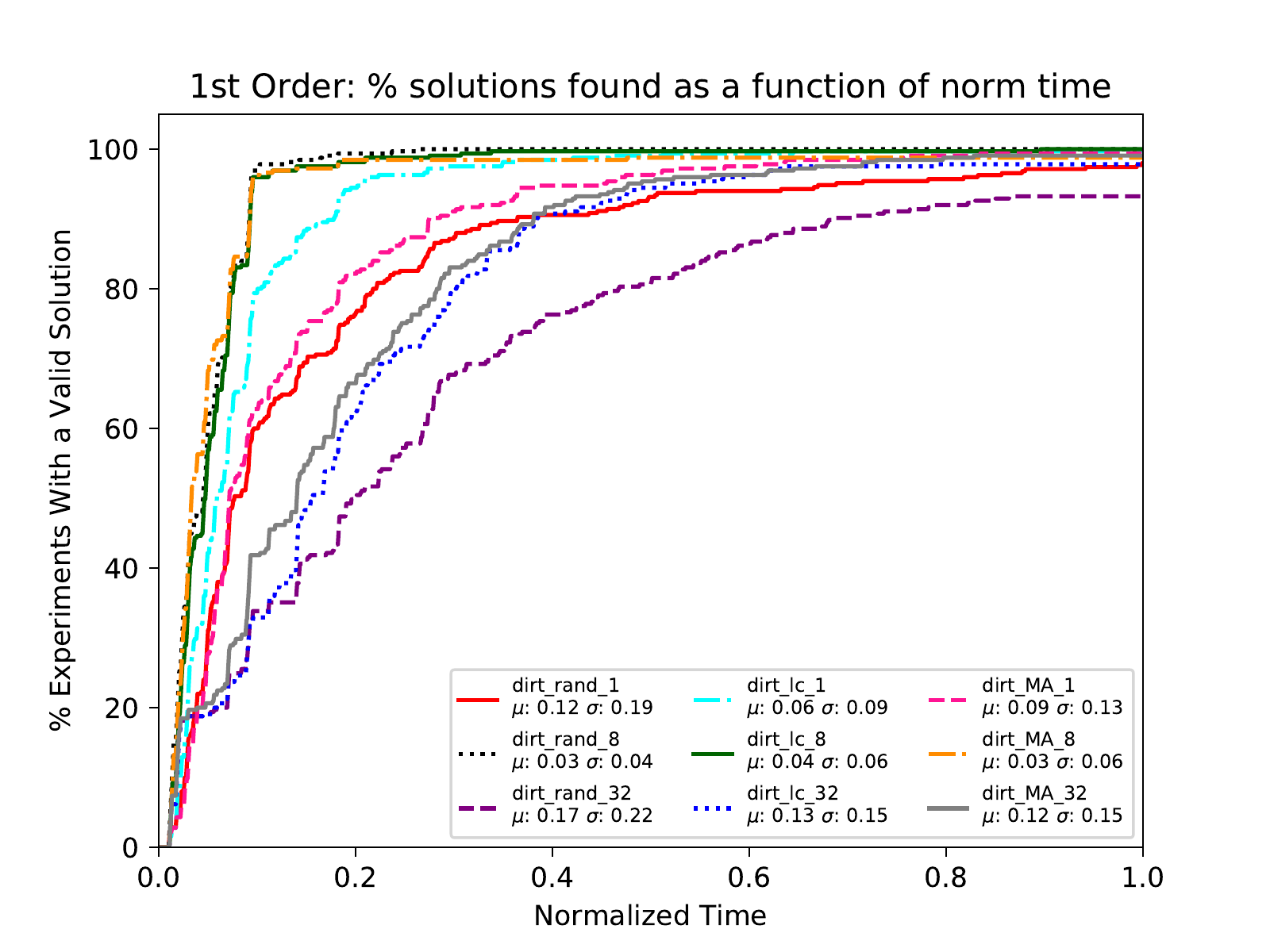}
         \includegraphics[width=.245\textwidth]{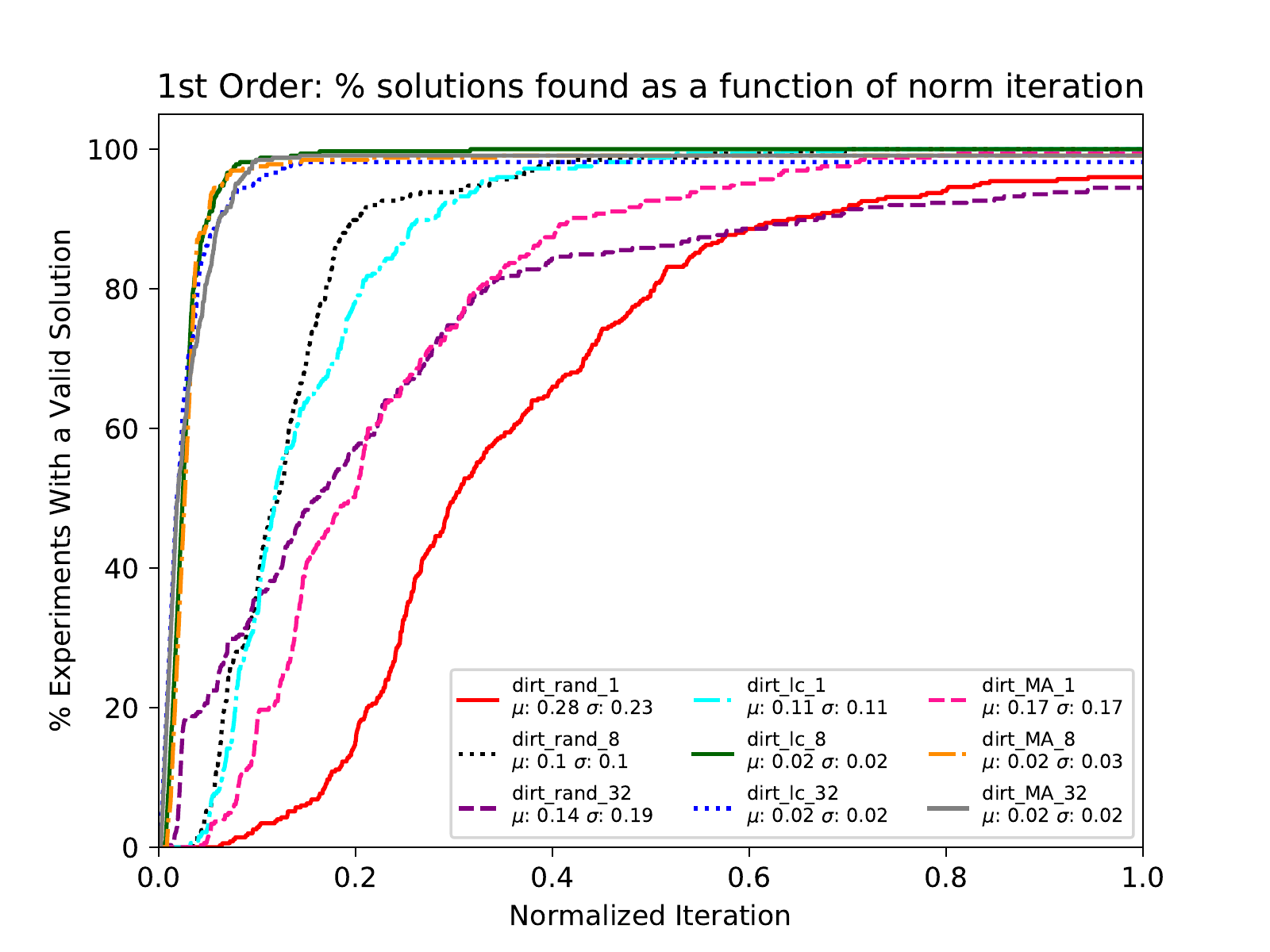}
         \vspace{-5mm}
         \caption{Results for first-order differential drive system in the city environments.}
         
         \vspace{-1mm}
         \label{fig:planner_analysis_fo}
\end{subfigure}\\
\begin{subfigure}[b]{\textwidth}
         \includegraphics[width=.245\textwidth]{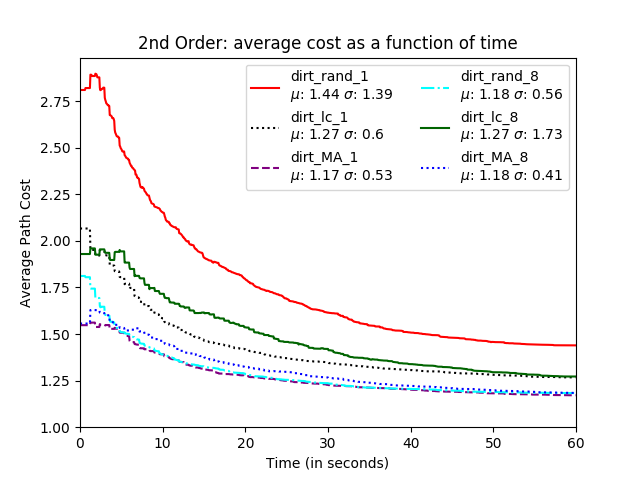}
         \includegraphics[width=.245\textwidth]{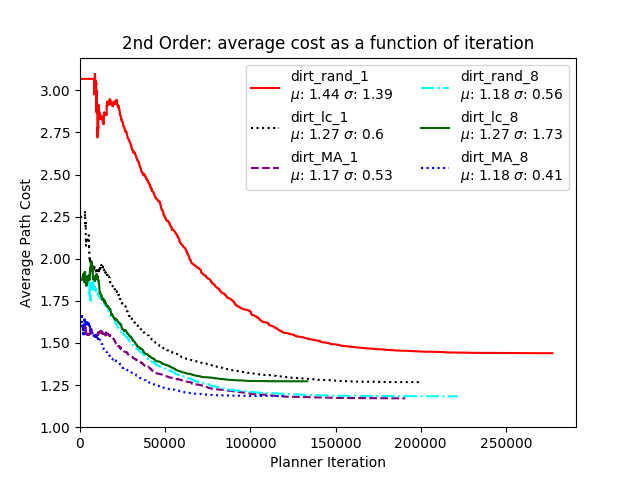}
         \includegraphics[width=.245\textwidth]{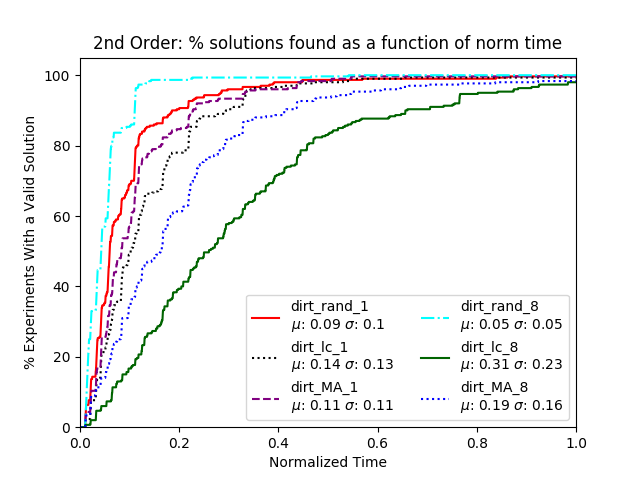}
         \includegraphics[width=.245\textwidth]{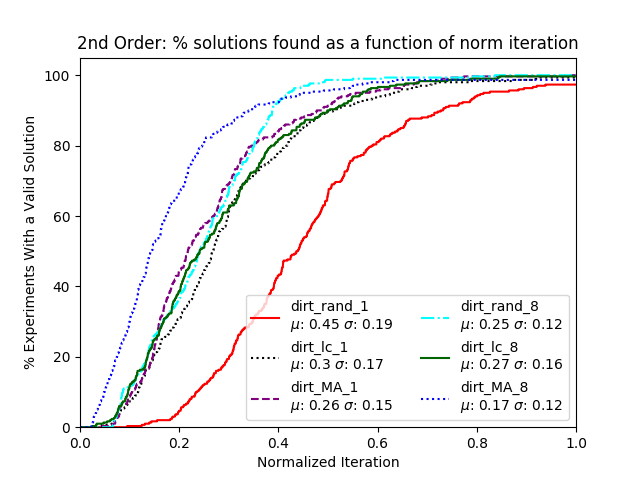}
         \caption{Results for second-order differential drive system in the city environments.}
         \label{fig:planner_analysis_so}
\end{subfigure}
    \vspace{-5mm}
\caption{For the left and middle-left figures, lower path cost is better. Path cost is normalized by dividing by the best path cost found for a  problem across any planner. For the right and middle-right figures, higher percentage of solutions found at an earlier normalized time/iteration is better. Time/iteration is normalized by dividing by the maximum time/iteration required by any planner to find a first solution.}
\end{figure*}
         

\begin{figure}[!h]
    \vspace{2mm}
    \centering
    \includegraphics[width=.14\textwidth]{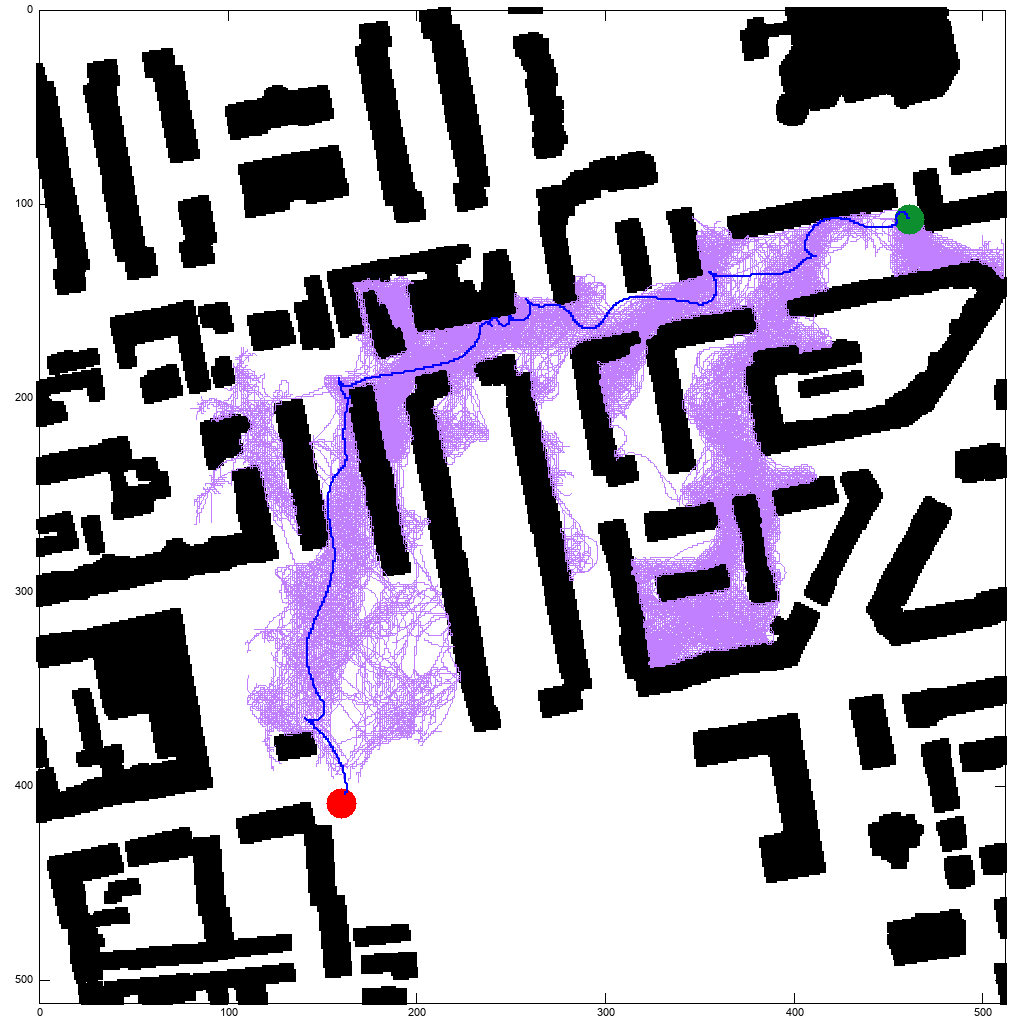}
    \includegraphics[width=.14\textwidth]{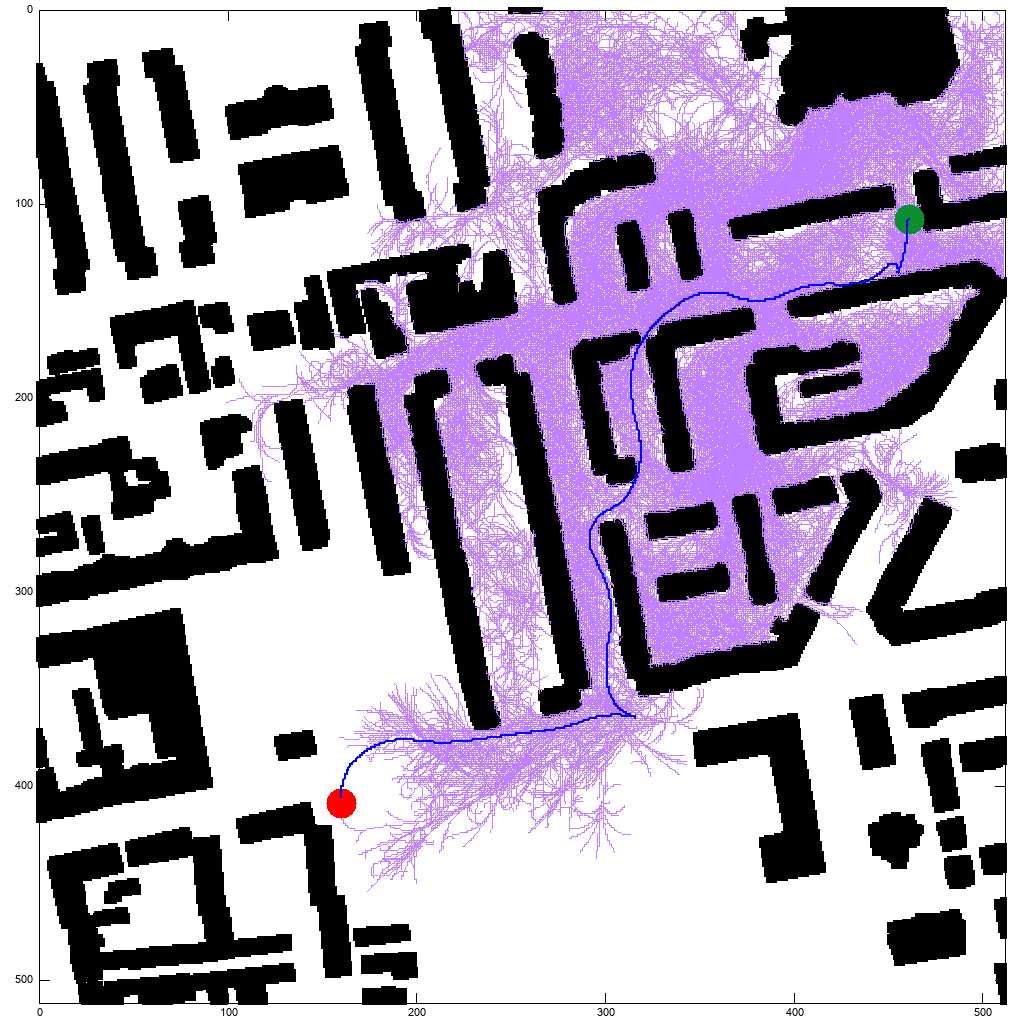}
    \includegraphics[width=.14\textwidth]{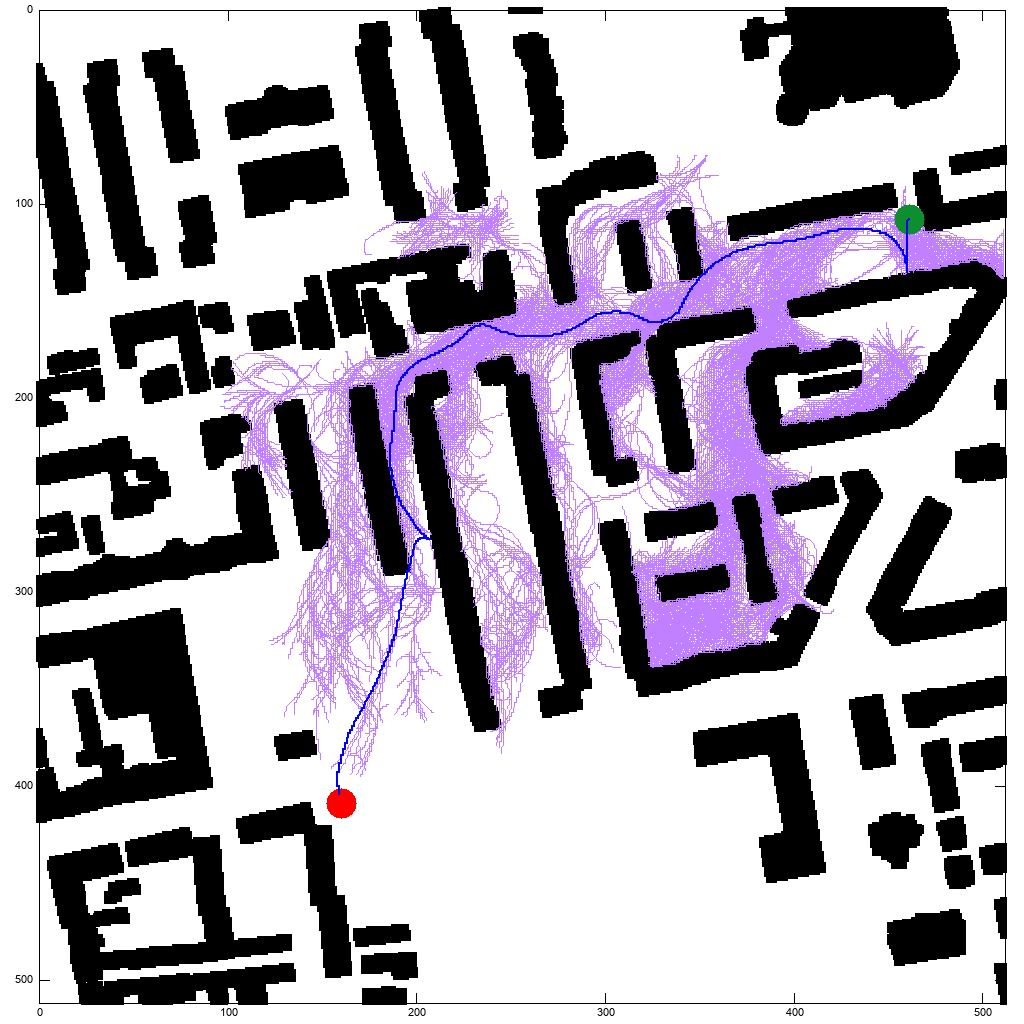}\\
    \includegraphics[width=.14\textwidth]{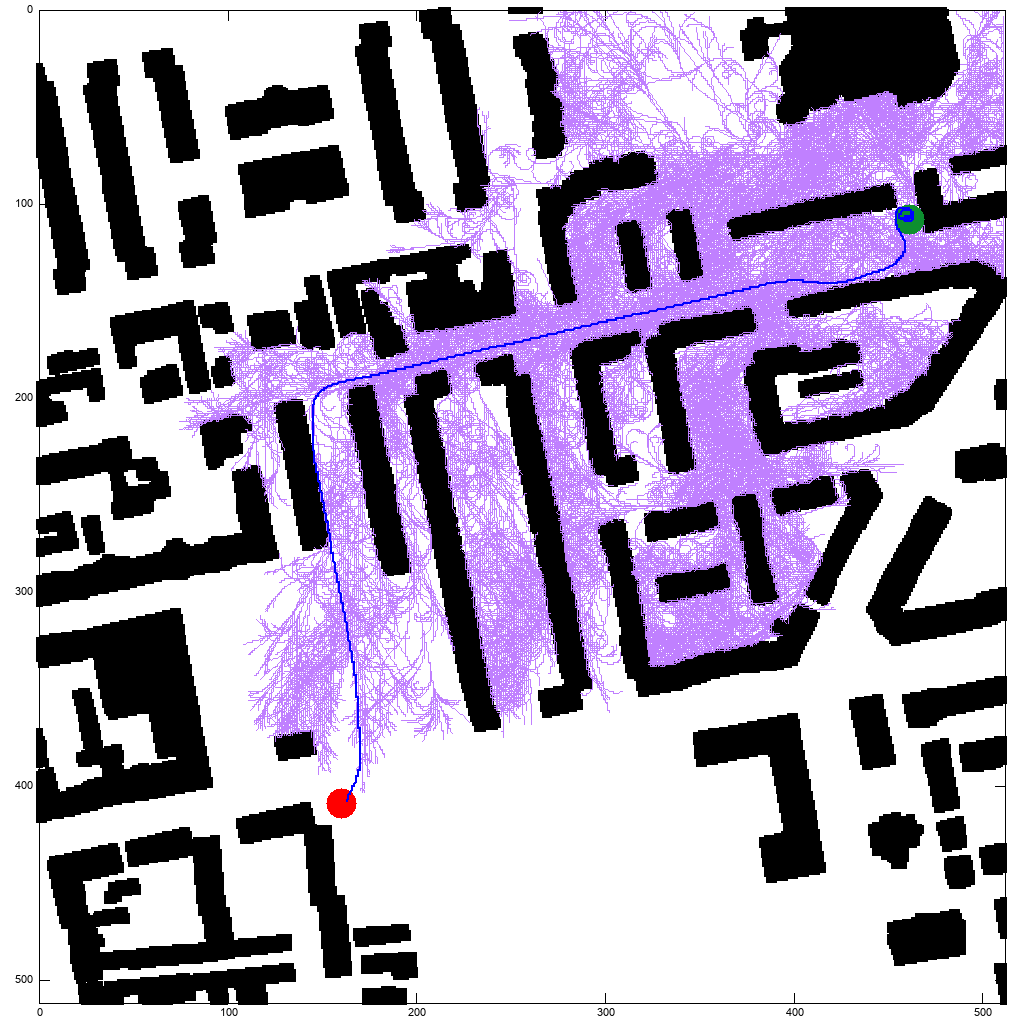}
    \includegraphics[width=.14\textwidth]{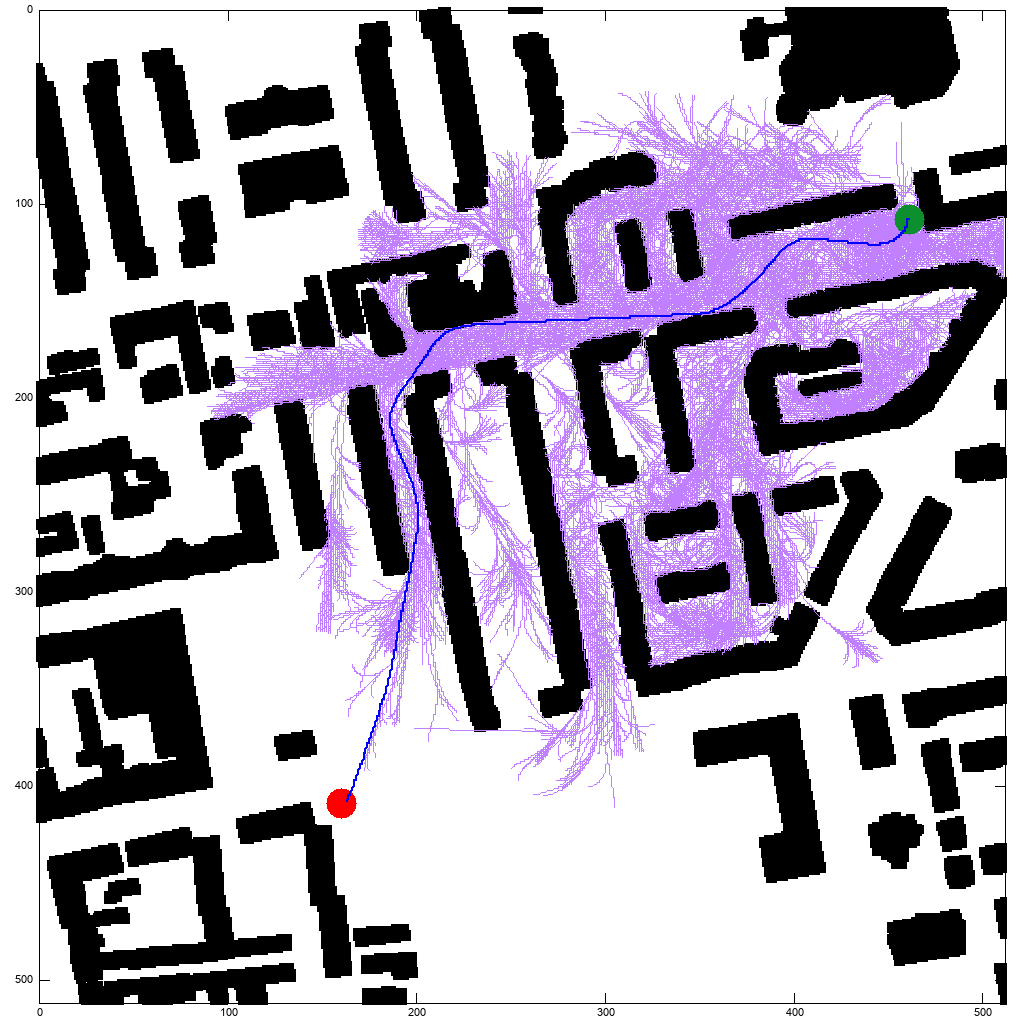}
    \includegraphics[width=.14\textwidth]{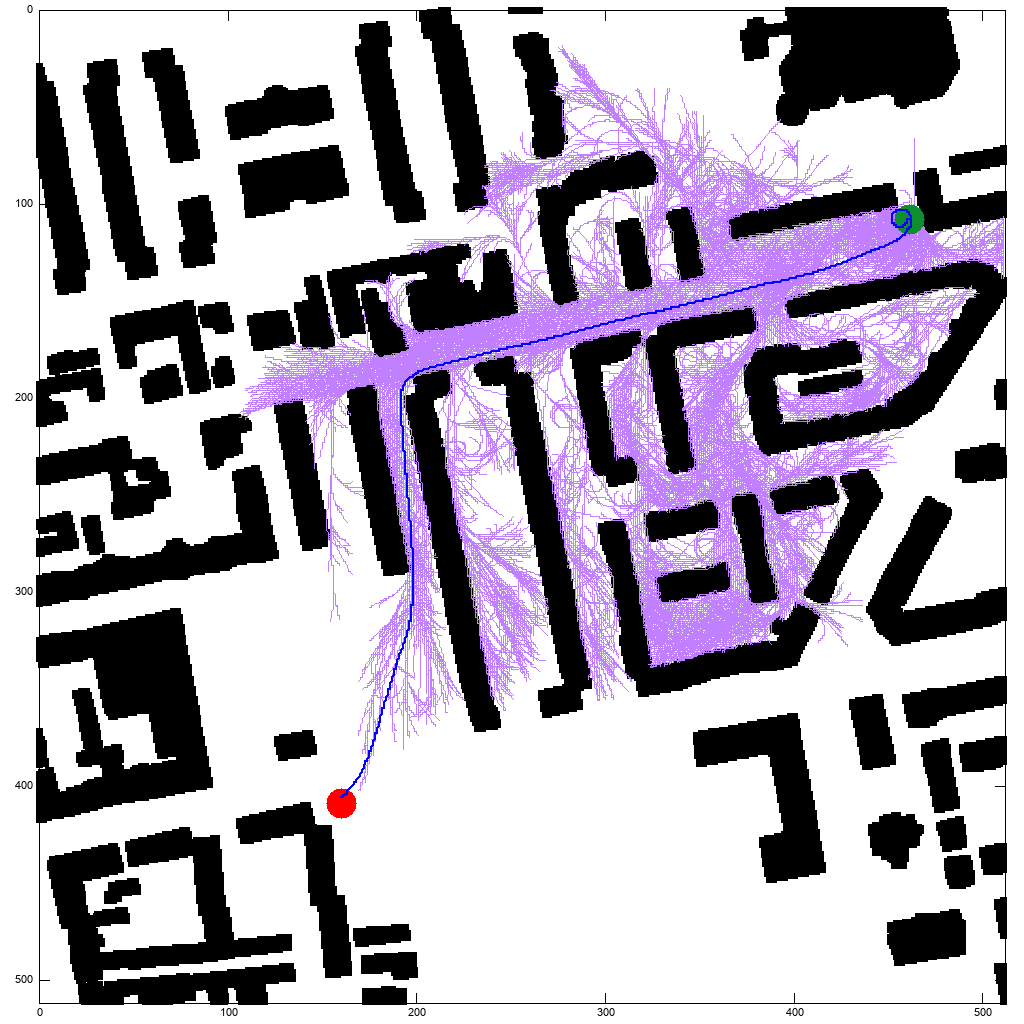}
    \vspace{-.5mm}
    \caption{Paths for the first-order system (top) from left to right are the random (blossom 32, cost 141), learned controller (lc) (blossom 8, cost 102), and learned controller with medial axis expansion (MA) (blossom 32, cost 103) and second order system (bottom) from left to right are the random (blossom 8, cost 101), lc (blossom 8, cost 82), and MA expansion (blossom 8, cost 84). Unlike the second-order learned controller, for the simpler, first-order system, the learned controller does not bias the controls forwards, leading to less smooth, higher cost solutions.}
    \label{fig:paths}
    \vspace{-1mm}
\end{figure}

\section{Results}\label{sec:results}
Experiments use a first order and a second order system, with the goal of minimizing the error of the learned controller and improve the performance of planning. We compare DIRT\cite{LB-DIRT} random with varying blossom numbers to DIRT using the learned controller and the proposed expansion functions. Smaller networks were chosen to reduce inference time at the expense of training error. Even though there is still training error for the learned controls, the learned controller reaches a target dispersion for predicted controls as shown in the corresponding plots and results in high-quality planning solutions.
\subsection{Data Collection Parameter Tuning}
Tuning the parameters for pruning influences the method's performance. The first-order system uses the following values: $\epsilon_E=0.1, \epsilon_R=\frac{\pi}{6}$. While for the second-order system: $\epsilon_E=0.2, \epsilon_R=\frac{\pi}{6}, \epsilon_V=\frac{v_{max}-v_{min}}{4}$. These parameters reduced the discretized set to 5\% of the original data size after pruning. Parameters were tuned to reduce the training error just below 0.01 combined control-duration and state space MSE for the first-order system and just below 0.05 control-duration MSE for the second-order system. These parameters result in the dataset visualized by the control data histogram in Figure \ref{fig:velDist}, which shows that the presented method favors bang-bang controls.

The proposed method preserves probabilistic completeness by ensuring that the learned controller has adequate coverage of the reachable space. While our results argue that the best performance is found by always calling the learned controller, these properties can be further assured by propagating one random control as one of the blossom controls.

The described method has a trade-off between the number of data points collected and the performance of the system. Our data collection method takes under 30 seconds to compare 2 million controls for the second-order system on a Ryzen 5 2600x processor before ultimately collection 20 thousand controls for training. Unlike related work that reduces the number of datapoints required for training \cite{quereshi2020motion}, the proposed method requires less computation point per datapoint.

\subsection{Model Training and Verification}
This section presents how the data set described in Section \ref{sec:dataset} is used to train a controller. Given a propagation time $t_{prop} \in [t_{min},t_{max}]$, the controller has to be effective for all possible valid state space changes $\delta x$, the \textit{reachable set}, that could arise by propagating a valid control in $\mathbb{U}$ from any valid state in $\mathbb{X}$.

 The first-order network architecture is ($\delta r, \delta \phi, \delta \theta$) $\rightarrow$ 64 $\rightarrow$ 16 $\rightarrow$ ($v_l, v_r, \delta t$) while the second order architecture is ($\delta r, \delta \phi, \delta \theta, v_l^{goal}, v_r^{goal}, v_l^{init}, v_r^{init}$) $\rightarrow$ 64 $\rightarrow$ 32 $\rightarrow$ ($u_l, u_r, \delta t$).  Training uses MSE as shown in Figure \ref{fig:training}.

\subsection{Application to Motion Planning}

 The planner was tested using the city benchmark environments \cite{sturtevant2012benchmarks} (figure \ref{fig:city_environments}). These environments were created by performing scans on real world cities, such as Berlin and Boston. Ten different planning problems were generated by sampling free start and goal states that have a minimum separation of $M$ pixels. Figure \ref{fig:roadmap} shows an example of a path in this environment.
 
 The following are compared: DIRT without learning (dirt\_rand), DIRT with a learned controller using sampling-based expansion (dirt\_lc), and DIRT with a learned controller using medial axis expansion (dirt\_MA). The blossom numbers used are 1, 8, and 32. Figures \ref{fig:planner_analysis_fo} and \ref{fig:planner_analysis_so} show the results of the first-order and second-order system respectively. The cost and solutions found are averaged and normalized: the best solution found by any planner has a cost/duration of 1. The planners were run for 60 seconds with 30 random seeds in 10 distinct planning problems in each of a number of different city environments (as shown in figure \ref{fig:city_environments}). Visualizations of the paths can be seen in Figure \ref{fig:paths}. 
 
 From the performance, the following is observed: the learned method is able to find low-cost solutions faster and in fewer iterations than traditional sampling-based motion planners. In fact, in the first-order system, the learned controller planners converge to lower cost solutions than the random planners. The first solution found with the learned controllers is higher quality than the first random solutions. The learned controller is able to find solutions in fewer iterations than DIRT random. Results show that the medial axis expansion is able to find lower cost solutions, in less time and fewer iterations than the sampling-based expansion method.

\section{Conclusion}\label{sec:conclusion}
This paper contributes a data efficient method to train a learned controller, which takes a state space difference vector as input and outputs a control and propagation duration. 
Our model can be used in sampling-based kinodynamic planning to find low-cost solutions quickly, while allowing to maintain asymptotic optimality properties. Unlike reinforcement learning methods, the proposed method creates high-quality, small supervised datasets, which can be quickly trained once for a robotic system. This work also proposed an exploratory and exploitative expansion function for kinodynamic planners to generate inputs for the learned controller. 

Since the resulting integration solves planning problems in fewer planning iterations, but can take more time per iteration due to inference costs, further improvement can be achieved by burning the learned network to an FPGA to reduce inference time. Moving forward, the expansion function may be learned in conjunction with the controller. The proposed methods can also be integrated with other machine learning primitives for planning, such as a learned sampling process. Next steps include to experiment on real vehicular system as well as extend the idea to more complex systems, such as legged systems. 

\bibliographystyle{format/IEEEtran}
\bibliography{bib/refs.bib}

\end{document}